\documentclass[a4paper,conference]{IEEEtran}
\usepackage[left=1.57cm,right=1.57cm,top=0.95cm,bottom=2.9cm]{geometry}
\IEEEoverridecommandlockouts
\usepackage{amsmath,amssymb,amsfonts}
\usepackage{algorithmic}
\usepackage{hyperref}
\usepackage{graphicx}
\usepackage{textcomp}
\usepackage[table,xcdraw]{xcolor}
\usepackage{csvsimple}
\usepackage{adjustbox}
\usepackage[T1]{fontenc}
\usepackage{cite}
\usepackage{multirow}
\def\BibTeX{{\rm B\kern-.05em{\sc i\kern-.025em b}\kern-.08em
    T\kern-.1667em\lower.7ex\hbox{E}\kern-.125emX}}
\begin{document}

\title{How important are socioeconomic factors for hurricane performance of power systems? An analysis of disparities through machine learning\\
%{\footnotesize \textsuperscript{*}Note: Sub-titles are not captured in Xplore and
%should not be used}
%\thanks{Identify applicable funding agency here. If none, delete this.}
}
\author{\IEEEauthorblockN{Alexys H. Rodríguez A.\IEEEauthorrefmark{1},
Abdollah Shafieezadeh\IEEEauthorrefmark{2} and Alper Yilmaz\IEEEauthorrefmark{3}}
\IEEEauthorblockA{Department of Civil, Environmental and Geodetic Engineering,
The Ohio State University\\
Columbus, Ohio 43210\\
Email: \IEEEauthorrefmark{1}rodriguezavellaneda.1@osu.edu,
\IEEEauthorrefmark{2}shafieezadeh.1@osu.edu,
\IEEEauthorrefmark{3}yilmaz.15@osu.edu}}
\maketitle

\begin{abstract}
This paper investigates whether socioeconomic factors are important for the hurricane performance of the electric power system in Florida. The investigation is performed using the Random Forest classifier with Mean Decrease of Accuracy (MDA) for measuring the importance of a set of factors that include hazard intensity, time to recovery from maximum impact, and socioeconomic characteristics of the affected population. The data set (at county scale) for this study includes socioeconomic variables from the 5-year American Community Survey (ACS), as well as wind velocities, and outage data of five hurricanes including Alberto and Michael in 2018, Dorian in 2019, and Eta and Isaias in 2020. The study shows that socioeconomic variables are considerably important for the system performance model. This indicates that social disparities may exist in the occurrence of power outages, which directly impact the resilience of communities and thus require immediate attention.
\end{abstract}


\section{Introduction}
Hurricane winds induce physical damage to power grid infrastructure and consequently outages that are often widespread. The outages may subsequently disrupt the services of other infrastructure systems, thus impacting communications, emergency services, and humanitarian aid in the critical times after the hazard. Aware of this complex environment, researchers have focused their efforts on understanding the physical vulnerabilities and assessing risks facing the power system, primarily from engineering perspective, to mitigate risks through optimal pre-allocation of resources \cite{dehghani2021intelligent, darestani2021life, salman2017multihazard} or post-disaster recovery plans \cite{dehghani2022multi, roman2019satellite, Kabir2019}.
\subsection{Social impacts of infrastructure disruptions}
Infrastructure systems have a strategic priority for the society \cite{NEHRP2008, Council2012, Standards2020}. However, the main requirements and guidelines that direct infrastructure design and engineering practices lack consideration of the socioeconomic impacts of infrastructure service disruptions \cite{Council2016}, which is deemed as a critical gap \cite{berkeley2010framework, Council2016}. Chang \cite{Chang2016} argues the importance of engineering, social science, interdisciplinary research, planning, and a growing amount of information sources for reducing infrastructure systems disruptions caused by natural hazards and for improving the resilience of certain socioeconomic vulnerable groups to disasters. 
\par
The socioeconomic impacts caused or exacerbated by infrastructure disruptions and their cascading effects during extreme events have been investigated by several multidisciplinary teams. For example, Mitsova et al. \cite{Mitsova2018} studied the electric power recovery times (dependent variable) in Florida caused by Hurricane Irma (2017). Analysis of data using spatial lag regression models at the county scale, and later on at the household scale \cite{Mitsova2019}, showed that racial minority groups with poor socioeconomic status are more impacted by power service disruptions.
\par
Environmental justice is a framework that focuses on socioeconomic differences between spatial units that are impacted and not impacted by environmental hazards. The analysis considers demographic (e.g. race, ethnicity, income or poverty, education) and/or environmental (distance to environmental hazards, others) indicators to analyze differential recovery rates in advance to natural disasters, therefore, facilitating identification of communities that are more vulnerable. Using this framework and quantitative regression analyses with three environmental justice vulnerability indexes (demographic, environmental hazard, environmental justice), and other indicator variables (terrain and distance to electric transmission lines), \cite{Sotolongo2021} found that specific vulnerable community groups in Puerto Rico had lower rates of recovery (restoration times) for their electric power system in the aftermath of Hurricane Maria (2017).
\par
\begin{figure}%[htbp]
\centerline{\includegraphics[width=1\columnwidth]{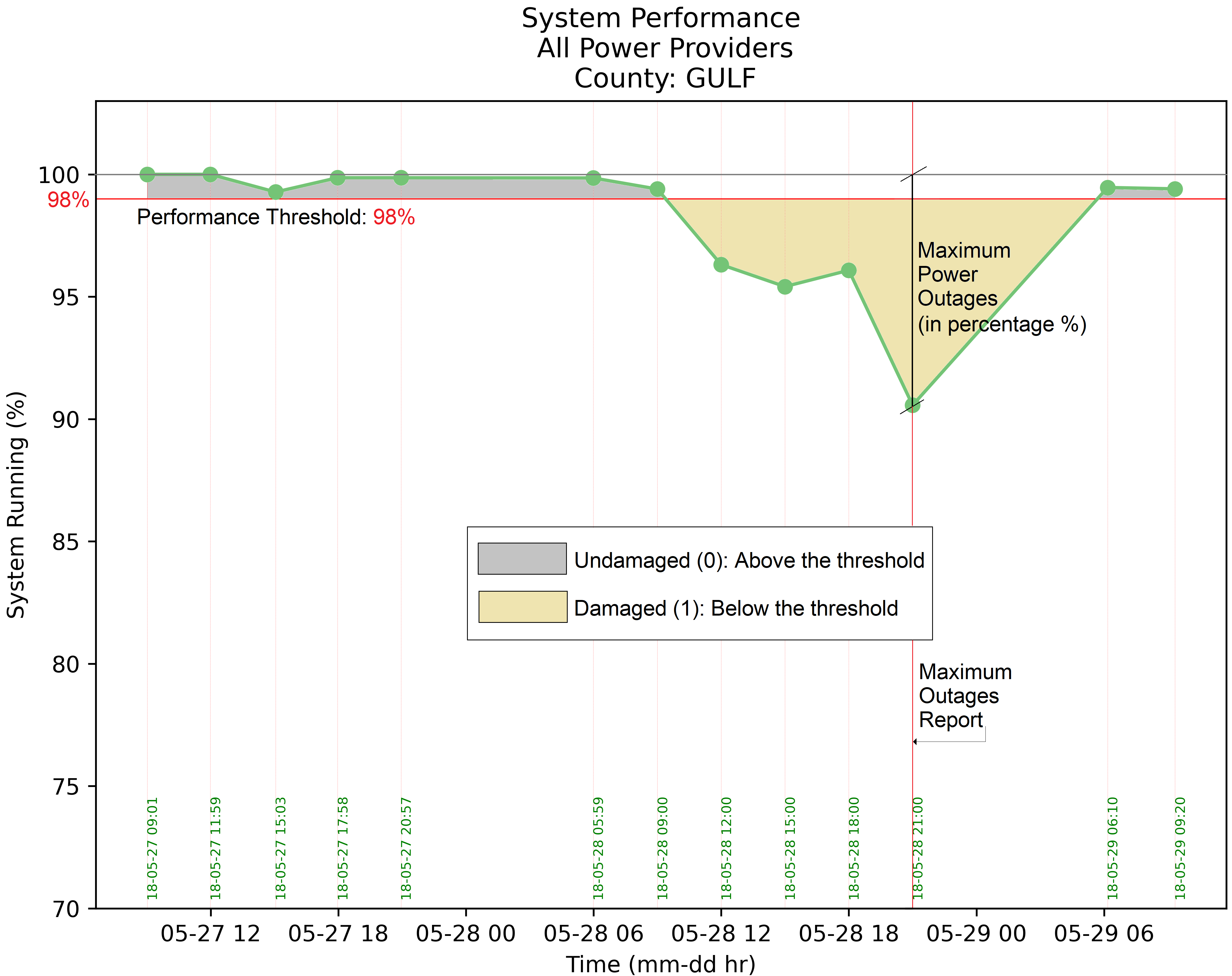}}
\caption{System Performance at outages threshold of 2\%. Damaged (1) and Undamaged (0) system at Maximum Power Outages (\%)}
\label{fig:systemperformace}
\end{figure}

Contributing to understanding the relationship between electricity outages and socioeconomic vulnerability, this paper presents a comprehensive study with four important qualities: (a) it analyzes the significance of socioeconomic variables using the Random Forest machine learning technique to classify outage performance, (b) it integrates data from multiple historical hurricanes in a single analysis, (c) it exclusively uses public data, and (d) it uses wind gust data to characterize the extreme wind hazard (velocities and times), see \ref{windhazard} ERA5 reanalysis data set. Studying the outage distribution at the county level in Florida in the aftermath of hurricane hazards, this study seeks to understand the impact of socioeconomic variables on \textbf{\textit{system performance}} that is represented by binary states of \textit{damaged system (1)} or \textit{undamaged system (0)}, when comparing the ``maximum power outages'' with the outage threshold of 2\% of the customers (see  in the Figure \ref{fig:systemperformace} the performance threshold of 98\%). In addition to socioeconomic variables, the model considers two groups of independent variables (see \eqref{eq}) including hurricane wind-related variables (current and past maximum wind speed and recovery times for each county, section \ref{windhazard}) and other variables including the number of customers and county urban/rural classification (see section \ref{othervariables}).
\begin{equation}
\begin{split}
\textit{system performance} \in \{damaged, undamaged\}& = \\
\textit{function(socioeconomic variables, wind hazard,}& \\ 
\textit{number of customers, urban/rural county)}&
\end{split}
\label{eq}
\end{equation}
% 
%\vspace{8pt}
% 
Given the system quality threshold of 98\%, the performance of the system can be measured in terms of maximum percentage of outages. Figure \ref{fig:systemperformace} shows the system performance for the Gulf County during Hurricane Alberto (2018). In this case, the \textit{x-axis}  shows the dates and times of 13 different damage reports, and the \textit{y-axis} shows the percentage of the system running. The maximum percentage of outages is reached when the system quality is at the lowest. For classification purposes using machine learning, the system performance can be considered to have two possible states, undamaged system (value of 0) for outages above the threshold, and damaged system (value of 1) for outages below the threshold.
\section{Data Set}
We studied the effect of five hurricanes including Alberto (2018), Michael (2018), Dorian (2019), Eta (2020), and Isaias (2020) in the State of Florida by considering the outages and socioeconomic parameters at county level.
\subsection{Outage Data}
The Florida Public Service Commission - PSC provides public power outages reports in PDF format for important wind events related to Florida, since 2018 \cite{por}. An automated data post-processing code is developed to generate a spatial database and extract the maximum number of outages for the hurricanes in the study. The original data source is summarized in Table ~\ref{tab:or}. The main characteristics of the reports, i.e. the number of reports, and the first and last time reported, vary notably across the hurricanes (see section \ref{dataintegration} for the explanation of outage data integration of five hurricane outage in one database).
\begin{table}[htb]\centering
\caption{Statistics of outage reports in PDF file format.}\label{tab:or}
\begin{tabular}{|l|c|l|l|}
\hline
\rowcolor[HTML]{EFEFEF} 
\multicolumn{1}{|c|}{\cellcolor[HTML]{EFEFEF}\textbf{Hurricane}} & \textbf{Reports} & \multicolumn{1}{c|}{\cellcolor[HTML]{EFEFEF}\textbf{First Report}} & \multicolumn{1}{c|}{\cellcolor[HTML]{EFEFEF}\textbf{Last Report}} \\ \hline \hline
Alberto                                                          & 13               & 5/27/2018 9:01                                                     & 5/29/2018 9:20                                                    \\ \hline
Dorian                                                           & 12               & 9/2/2019 18:51                                                     & 9/4/2019 18:00                                                    \\ \hline
Eta                                                              & 15               & 11/8/2020 21:00                                                    & 11/12/2020 18:00                                                  \\ \hline
Isaias                                                           & 6                & 8/1/2020 18:00                                                     & 8/2/2020 18:00                                                    \\ \hline
Michael                                                          & 131              & 10/9/2018 21:00                                                    & 11/6/2018 11:18                                                   \\ \hline
\end{tabular}
\end{table}
\subsection{Socioeconomic Variables}
Socioeconomic county indicators that are traditionally relevant in environmental justice frameworks and infrastructure disruption studies are considered here. These indicators relate to race/ethnicity, general population profile, population dependence, housing and households, educational attainment, income, and poverty. Table ~\ref{tab:se} shows detailed information about the socioeconomic variables used in this study, coming from the 5-year American Community Survey \cite{acs5} for the previous year to each hurricane landfall. Therefore, ACS-5 2017, 2018, and 2019 are considered in this study. 
\begin{table}%[htb]
\centering
\caption{Socioeconomic variables}\label{tab:se}
\begin{adjustbox}{width=\columnwidth,center}
\begin{tabular}{|c|l|c|}
\hline
\rowcolor[HTML]{EFEFEF} 
\textbf{Type}                                & \multicolumn{1}{c|}{\cellcolor[HTML]{EFEFEF}\textbf{Variable}} & \textbf{Id.}            \\ \hline \hline
                                             & \% population White                                            & 2                       \\ \cline{2-3} 
                                             & \% Latino                                                      & 3                       \\ \cline{2-3} 
                                             & \% African                                                     & 4                       \\ \cline{2-3} 
                                             & \% Asian                                                       & 5                       \\ \cline{2-3} 
                                             & \% Indian                                                      & 6                       \\ \cline{2-3} 
\multirow{-6}{*}{Race/Ethnicity}             & \% Other                                                       & 7                       \\ \hline
                                             & Population                                                     & 8                       \\ \cline{2-3} 
\multirow{-2}{*}{General Population Profile} & Population Density                                             & 9                       \\ \hline
                                             & \% population < 5 years                                        & 10                      \\ \cline{2-3} 
                                             & \% population > 65 years                                       & 11                      \\ \cline{2-3} 
                                             & \% population no vehicle                                       & 12                      \\ \cline{2-3} 
                                             & \% population public assistance                                & 13                      \\ \cline{2-3} 
                                             & \% population limited English                                  & 14                      \\ \cline{2-3} 
                                             & \% population disability                                       & 15                      \\ \cline{2-3} 
\multirow{-7}{*}{Dependence}                 & \% population health insurance                                 & 16                      \\ \hline
                                             & \% renter occupied housing                                     & 17                      \\ \cline{2-3} 
                                             & \% income for renting > 30\%                                   & 18                      \\ \cline{2-3} 
                                             & \% single > 65 years (rented)                                  & 19                      \\ \cline{2-3} 
\multirow{-4}{*}{Housing \& Households}      & \% single > 65 years (owned)                                   & \multicolumn{1}{l|}{20} \\ \hline
Educational Attainment                       & \% < High School                                               & 21                      \\ \hline
                                             & Unemployment rate                                              & 22                      \\ \cline{2-3} 
                                             & \% < poverty level                                             & 23                      \\ \cline{2-3} 
\multirow{-3}{*}{Income \& Poverty}          & \% income for renting > 30\%                                   & 18                      \\ \hline
\end{tabular}
\end{adjustbox}
\end{table}
\subsection{Wind Hazard} \label{windhazard}
\begin{figure}%[htbp]
\centerline{\includegraphics[width=1\columnwidth]{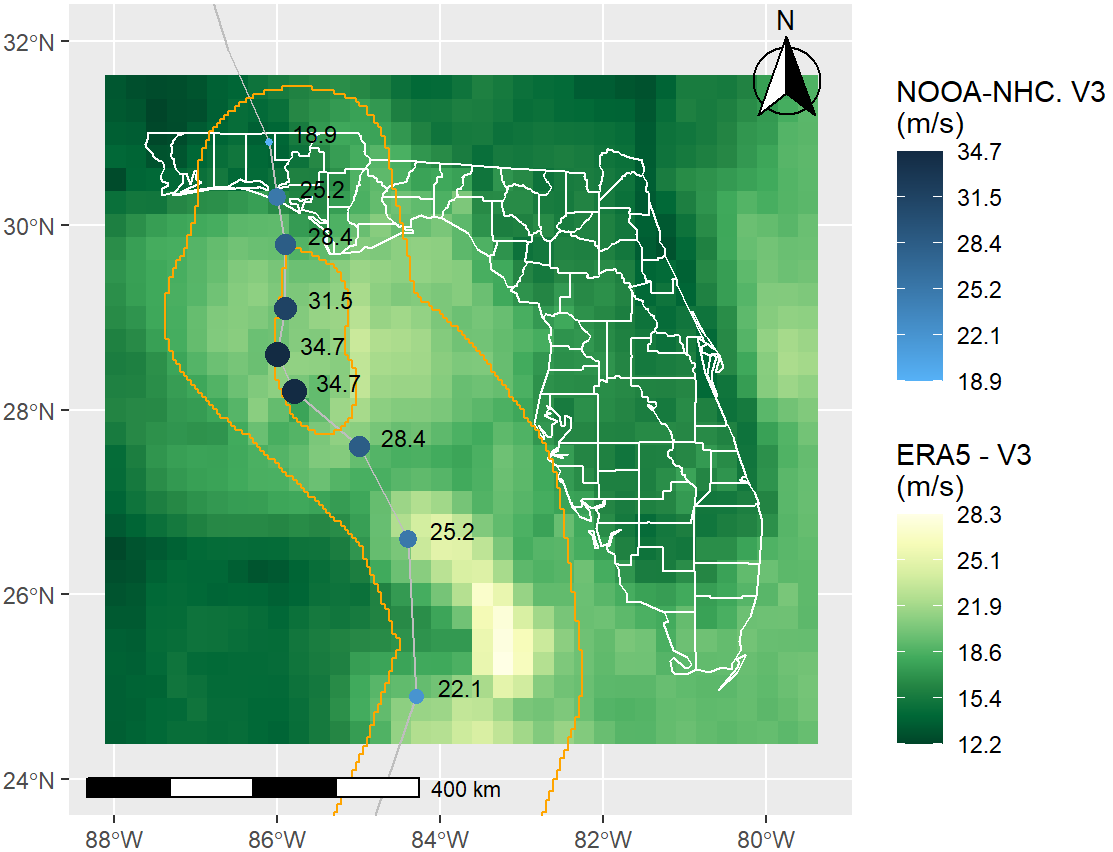}}
\caption{ERA5 maximum wind gust Vs. NOOA-NHC best track. Alberto Hurricane. The best track is represented by the grey line with some intensity measures (points) along it, ranging from 18.92 (cyan) to 34.68 (dark blue) m/s.}
\label{fig:era5}
\end{figure}

The European Centre for Medium-Range Weather Forecasts (ECMWF) provides the data set ERA5, which among many climate variables (land, atmospheric, and oceanic), supplies wind gust (\textbf{\textit{fg10}}) velocities. ERA5 reanalysis data set reconstruct past weather and climate based on  model information (past short-range weather forecast) combined with observations, i.e. data assimilation. The observations includes field measurements and satellite information from multiple sources and scales. The spatial resolution of ERA5 is a grid of 0.25 decimal degrees covering the entire world surface at different pressure levels, with hourly information from 1979 to the present. The variable \textbf{\textit{fg10}} is available in the data set ``ERA5 hourly data on single levels'' and represents 3-seconds wind gusts ($V_3$) in m/s, at 10 meters anemometer height, and under open space conditions \cite{hersbach2020era5}. For Hurricane Alberto, in Figure \ref{fig:era5}, the maximum $V_3$ from ERA5 is compared against the best track information provided by the National Hurricane Center (NHC) - National Oceanic and Atmospheric Administration (NOAA)\footnotemark{}. In order to compare, the intensity $V_{60}$ in knots provided by NHC-NOAA was transformed to $V_3$ in m/s using \cite{Durst1960}.
\footnotetext{NHC-NOAA maintains the Hurricane Databases (HURDAT) and provides best track files for multiple historical storms, including 6-hours interval estimations of 60-seconds ($V_{60}$) maximum sustained surface winds in knots. (\url{https://www.nhc.noaa.gov/gis/})}
From the hourly time history in the time frame of each hurricane outage report (see Table ~\ref{tab:or}), a spatiotemporal analysis was performed to extract wind hazard relate features including (a) the maximum $V_3$ velocity reported for the ERA5 pixels intersecting each county area (maximum in space) at the time of each outage report and (b) the past maximum $V_3$ velocity reported for the pixels time history intersecting each county area (maximum in space and in time). The maximum velocity in (a) represents the wind speed at the moment of each outage report, and the one in (b) represents the maximum wind speed faced by each county during the hurricane time frame previous to each outage report. The time in days from the event (b) to each outage report provides an additional feature that represents (c) the amount of time that has been available to each county to recover. For this analysis, these three features corresponding to the ``maximum outage report'' are considered.

\subsection{Other Variables} \label{othervariables}
Two additional independent variables are included in the model: (a) urban (1) / rural (0) county classification and (b) the number of power utility customers per county.
\subsection{Database Integration} \label{dataintegration}
An integrated spatial database is created with all the feature information related to the five hurricanes' ``maximum outage report'', all the socioeconomic variables (22), the hazard features (3), urban/rural classification, and the number of customers, at the county scale. Table ~\ref{tab:vs} shows a summary of the variables. The total number of columns is 28 (1 dependent - 27 independent). The Id column is used later in this document to identify (label) variables in the results graphs.
\begin{table}%[htb]
\centering
\caption{Variables Summary}\label{tab:vs}
\begin{adjustbox}{width=\columnwidth,center}
\begin{tabular}{|l|l|l|c|}
\hline
\rowcolor[HTML]{EFEFEF} 
\multicolumn{1}{|c|}{\cellcolor[HTML]{EFEFEF}\textbf{Model}} & \multicolumn{1}{c|}{\cellcolor[HTML]{EFEFEF}\textbf{Type}} & \multicolumn{1}{c|}{\cellcolor[HTML]{EFEFEF}\textbf{Variable}} & \textbf{Id.} \\ \hline \hline
Dependent                                                    & System performance                                         & Damaged (1) - undamaged (0)                                    & 1            \\ \hline
Independent                                                  & Socioeconomic                                              & Table \ref{tab:se}                                             & 2-23      \\ \hline
Independent                                                  & Other variables                                            & Urban (1) - Rural (0)                                          & 24           \\ \hline
Independent                                                  & Other variables                                            & Number of customers                                            & 25           \\ \hline
Independent                                                  & Wind hazard                                                & Current maximum velocity                                       & 26           \\ \hline
Independent                                                  & Wind hazard                                                & Past maximum velocity                                          & 27           \\ \hline
Independent                                                  & Wind hazard                                                & Recovery time from maximum impact                           & 28           \\ \hline
\end{tabular}
\end{adjustbox}
\end{table}

\section{Random Forest}

Random Forest, developed by Breiman \cite{breiman2001random}, is an ensemble of many decision trees from different bootstrap samples (bagging), but with de-correlated trees. It can capture the general (linear or non-linear) structure of a data set (low bias), and it can be robust to outliers and noise. Handling binary, categorical, and numerical features without re-scaling or transforming, the implementation of Random Forest is simple, flexible, without exigent parameter tuning, and with low computational cost. It can also be used for regression, classification, or survival analysis in high dimensional data. For more details about the method with code examples, see Chapter 8 of \cite{james2013introduction}.

\subsection{Variable Importance} \label{variableimportance}

Node impurity is a measure of the uncertainty or diversity (non-homogeneity) of the labels at each tree node. When the labels inside a node belong to the same class the impurity is zero, and on the contrary, high impurity is related to mixed labels in a node. The most important feature with higher impurity reduction when splitting will be located as the root node (top node of the inverted tree) resulting in more pure children nodes (groups of majority class are isolated) compared with parent nodes. Gini index and Entropy are similar concepts directly related to the degree of inhomogeneity of the samples inside a node. The Gini index measures the probability of picking two distinct classes inside the node, being directly proportional to node impurity, and similarly, the Entropy is the expected self-information, a measure of uncertainty, or disorder of a sample set contained in a node. In Random Forest, the most important attributes can be selected, considering maximum information gain when splitting, a technique called Mean Decrease Impurity (MDI), but this approach also considers noisy features as important \cite{li2019debiased} and it is biased towards (a) categorical variables with multiple classes, (b) continuous variables, or (c) variables with high category frequencies \cite{strobl2007bias, nembrini2018revival, nicodemus2011stability}. A preferred method that overcomes \cite {ziegler2014mining} the deficiencies of MDI is permutation importance, also known as Mean Decrease of Accuracy (MDA), but it requires extra computing cost. With MDA an important variable has a considerable decrease (big difference) in the model accuracy comparing its Out-of-Bag (OOB) prediction error (model including original variable) with the new  OOB average prediction error obtained from performing multiple random permutations (shuffles) of the variable content inside the model \cite{breiman2001random}, and the process is repeated independently for each variable in the model. The mentioned difference is the variable permutation importance, therefore, MDA measures how the score decreases when the variable is not available. Neither MDI nor MDA reflect the predictive significance of a feature, but its relative importance for the specific model \cite{pedregosa2011scikit}.

\section{Experiments}

From the 335 county records in the integrated database (see \ref{dataintegration}), only 11\% (37 samples) of the reported outage percentages correspond to ``damaged'' system performance status (using an outage threshold of 2\%), and the remaining 89\% (298 samples) correspond to ``undamaged''. The implementation of Random Forest was performed 1000 times, each one with a different testing-training (30\%-70\%) random splitting (see section \ref{traintest} below), variable importance analysis, and optimized search o hyper-parameters through k-fold cross-validation, using HalvingRandomSearchCV Scikit-Learn (Machine Learning in Python) \cite{pedregosa2011scikit}. 

\subsection{Training and testing data sets} \label{traintest}

In machine learning supervised classification, the training and testing sets are assumed to have the same distribution \cite{Coleman2021}; hence, the random partitioning in training (70\%) and testing (30\%), and the whole Random Forest analysis was performed 1000 thousand times to verify similar model results across different runs.

\subsection{Tuning Hyper-parameters} \label{randomhyper}

The selection of a optimum set of hyper-parameters is based on increasing the performance of random search using successive halving \cite{jamieson2016non, li2017hyperband}, based on cyclic evaluation of parameters using few system resources at the first try, then filtering top-scoring candidate across iterations, and finally, in the last iteration, using an exhaustive search in few optimal parameters.

\subsection{Model performance}

The \textbf{\textit{relevant}} elements are those counties reporting a ``damaged'' (positive class) system, and after running Random Forest, the \textbf{\textit{selected}} elements are counties classified in the same positive class.

Class-specific evaluation of the model can be analyzed using sensitivity, specificity, precision, recall, and f1-score. The trade-off between sensitivity and specificity, defined by a probability threshold of belonging to the positive or negative class, can be seen in The Receiver Operating Characteristics (ROC) and the corresponding Area Under the Curve (AUC). A good classification model for all the classes has an AUC close to one (1), the maximum value. For a detailed compendium of this evaluation metrics in Machine Learning, see Chapter 2 of \cite{zheng2015evaluating}.

\section{Results and Discussion} \label{results}

The optimum sets of hyper-parameters using random search successive halving for the mean model are (a) criterion: entropy, (b) maximum number of tree nodes: 10, (c) maximum number of features per tree: 4 out of 27, (d) number of trees in the forest: 10, and (e) the minimum number of samples to split a node: 7.

\begin{figure}%[htbp]
\centerline{\includegraphics[width=6cm]{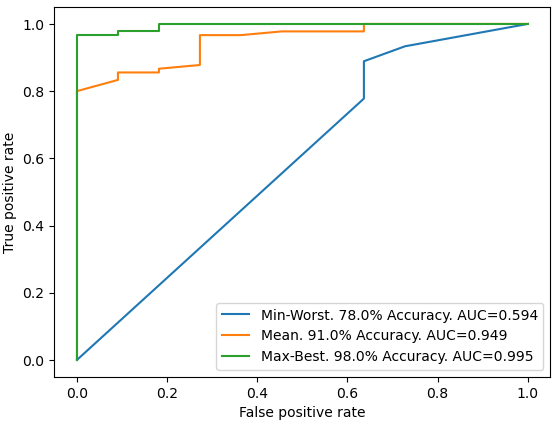}}
\caption{ROC Curve - AUC for worst, average, and best model. Different PDF for the 1000 random testing and training partitions.}
\label{fig:roc}
\end{figure}

\begin{figure}%[htbp]
\centerline{\includegraphics[width=6cm]{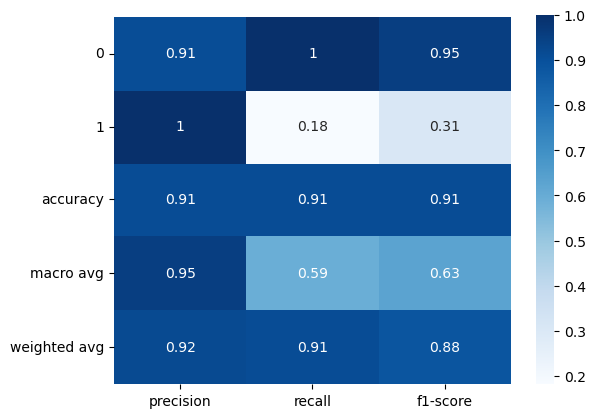}}
\caption{Classification report (mean model). The global accuracy of the model (91\%) contrasts with the low quality of the model for the ``damaged'' class, with an f1-score of 31\%}
\label{fig:report}
\end{figure}

The minimum (worst) and the maximum (best) model accuracies across 1000 iterations are 78\% and 98\%, respectively. These different results (maximum difference of 20\%), indicate a non similar probability distribution function (PDF) for all the random training and testing partitions, more so considering that the average model accuracy is 91\%. Figure \ref{fig:roc} shows that all the 1000 implemented models are different ranging from really bad performance, the min model with AUC of 0.594, to  excellent performance with the best model having AUC of 0.995. Despite the average model having a good AUC value of 0.949, Figure \ref{fig:report} shows that this model cannot detect the ``damaged'' class (1) very well due to the low f1-score of 31\% (high precision of 100\% and low recall of 18\%), but is highly trustable when it does. In Figure \ref{fig:report}, the results confirm really good performance of the mean model for the ``undamaged'' class (0).

\begin{figure}%[htbp]
\centerline{\includegraphics[width=1\columnwidth]{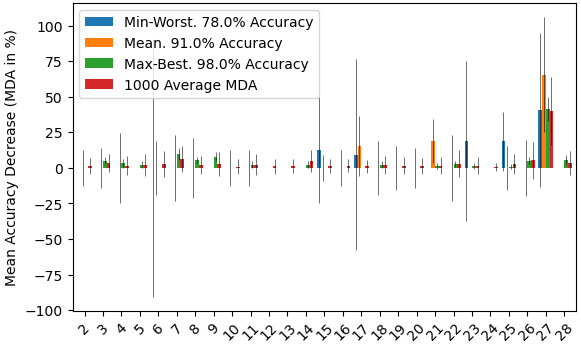}}
\caption{Permutation importance MDA. The most important variable is ``past maximum velocity (27)''. Different MDA across the 1000 models. High standard deviations (vertical black thin line). Unbalanced data set issues.}
\label{fig:mda}
\end{figure}

The results for variable importance are analyzed for the three representative models, namely the best (maximum), the worst (minimum), and the mean model, in addition to the average importance across the 1000 models. It should be noted that MDA addresses the MDI bias (see section \ref{variableimportance}) by removing wrongly selected variables or reducing the importance of others. Therefore, the analyses here are focused on MDA results, whose importance in percentage are ordered in Table \ref{tab:mda}. Each variable in each model has importance in percentage (from 0\% to 100\%) represented by the bars height in the Figure \ref{fig:mda}, where the standard deviation is depicted as a thin vertical line, or by a number in Table \ref{tab:mda}, which in addition has the variable importance average (not only the representative models, but the 1000 repetitions). Note that the worst model (min) can have higher importance for the same variable, in comparison with the best model (max).

The analysis using MDA reveals that the wind hazard variables represent the 49\% of the importance, including the \textit{past maximum velocity} with an average score of 39.81\%, the \textit{current maximum velocity} with 5.42\%, and the recovery time (days) each county had from the time of highest impact (stronger wind) to the time of maximum outages report, with 3.5\%. The top-five socioeconomic variables, which represent the 48\% of the model importance, are (a) percentage of Other population (Id = 7) with 6.31\%, (b) percentage of the population with Limited English speaking ability (Id = 14) with 4.99\%, (c) percentage of the Latino population (Id = 3) with 3.35\%, (d) Unemployment Rate (Id = 22) with 2.91\%, and (e) percentage of the Indian-American population (Id = 6) with 2.77\%.

\begin{table}%[htb]
\centering
\caption{Permutation importance MDA (in percentage).}\label{tab:mda}
\begin{adjustbox}{width=\columnwidth,center}
\begin{tabular}{|c|c|c|c|c|c|}
\hline
\rowcolor[HTML]{EFEFEF} 
\textbf{Id.} & \textbf{Min-Worst 78\%} & \textbf{Mean 91 \%} & \textbf{Max-Best 98\%} & \textbf{Avg. MDA} & \textbf{Avg. MDA-Std} \\ \hline \hline
\textit{27}  & 40.62                   & 65.62               & 41.31                  & 39.81             & 23.89                 \\ \hline
\textit{7}   & 0                       & 0                   & 9.86                   & 6.31              & 9.44                  \\ \hline
\textit{26}  & 0                       & 0                   & 5.16                   & 5.42              & 13                    \\ \hline
\textit{14}  & 0                       & 0                   & 2.35                   & 4.99              & 7.75                  \\ \hline
\textit{28}  & 0                       & 0                   & 5.63                   & 3.5               & 8.74                  \\ \hline
\textit{3}   & 0                       & 0                   & 5.16                   & 3.35              & 6.45                  \\ \hline
\textit{22}  & 0                       & 0                   & 2.82                   & 2.91              & 9.45                  \\ \hline
\textit{6}   & 0                       & 0                   & 0                      & 2.77              & 9.2                   \\ \hline
\textit{9}   & 0                       & 0                   & 7.98                   & 2.73              & 8.71                  \\ \hline
\textit{25}  & 18.75                   & 0                   & 0.94                   & 2.68              & 7                     \\ \hline
\textit{11}  & 0                       & 0                   & 2.35                   & 2.31              & 7.23                  \\ \hline
\textit{5}   & 0                       & 0                   & 2.35                   & 2.28              & 7.63                  \\ \hline
\textit{18}  & 0                       & 0                   & 2.35                   & 2.2               & 6.24                  \\ \hline
\textit{8}   & 0                       & 0                   & 5.63                   & 1.98              & 6.52                  \\ \hline
\textit{23}  & 18.75                   & 0                   & 1.41                   & 1.68              & 6.09                  \\ \hline
\textit{4}   & 0                       & 0                   & 3.29                   & 1.66              & 6.82                  \\ \hline
\textit{19}  & 0                       & 0                   & 0                      & 1.57              & 5.87                  \\ \hline
\textit{21}  & 0                       & 18.75               & 1.41                   & 1.56              & 5.99                  \\ \hline
\textit{16}  & 0                       & 0                   & 0                      & 1.52              & 4.52                  \\ \hline
\textit{2}   & 0                       & 0                   & 0                      & 1.36              & 5.78                  \\ \hline
\textit{12}  & 0                       & 0                   & 0                      & 1.31              & 5.26                  \\ \hline
\textit{15}  & 12.5                    & 0                   & 0                      & 1.24              & 4.78                  \\ \hline
\textit{13}  & 0                       & 0                   & 0                      & 1.23              & 4.82                  \\ \hline
\textit{20}  & 0                       & 0                   & 0                      & 1.12              & 5.69                  \\ \hline
\textit{17}  & 9.38                    & 15.62               & 0                      & 1.09              & 4.36                  \\ \hline
\textit{10}  & 0                       & 0                   & 0                      & 0.91              & 5.36                  \\ \hline
\textit{24}  & 0                       & 0                   & 0                      & 0.52              & 2.69                  \\ \hline
\end{tabular}
\end{adjustbox}
\end{table}

\section*{Conclusion}

Socioeconomic variables are found to be considerably important for the outage performance of electrical power system in Florida against hurricanes. Results indicate that the social disparities are represented mainly by race/ethnicity (percentage of the Other, Latino, and Indian-American population), income and poverty (unemployment rate), and dependence (percentage of the population with limited English speaking ability). These relevant socioeconomic conditions should be considered when making resilience decisions for future wind hazards in the study area. Nonetheless, more studies need to be conducted to reach a sound conclusive remarks due to the fact that (a) the issue of unbalanced data set may cause variation between training and testing probability distributions across the repetitive experiments, and (b) additional data of hurricane events may need to be included in the database to increase the fidelity of the models.

\vspace{12pt}
% \printbibliography %Prints bibliography
\bibliography{main}
\bibliographystyle{ieeetr}
\end{document}